\documentclass{applemlr}
\usepackage{amsmath}
\usepackage{enumerate} 
\usepackage{algorithm}
\usepackage{algpseudocode}
\usepackage{amsfonts}
\usepackage{amsthm}
\usepackage{newtxtt}
\usepackage{cleveref}
\usepackage{diagbox} 
\usepackage{colortbl}
\usepackage{bbm}
\usepackage{amssymb}
\usepackage{xspace}
\usepackage{wrapfig}
\usepackage{adjustbox}
\usepackage{tabularx}
\usepackage{booktabs}
\usepackage{mathtools}
\usepackage{tikz}
\usepackage{enumitem}
\usepackage{silence}
\usepackage{dsfont}
\usepackage[table]{xcolor}
\usepackage[dvipsnames]{xcolor}
\usepackage{multirow}
\usepackage{makecell}
\usepackage{xfakebold}
\usepackage{amsmath,amsfonts,bm}

\def\eqref#1{equation~\ref{#1}}
\def\1{\bm{1}}

\DeclareMathAlphabet{\mathsfit}{\encodingdefault}{\sfdefault}{m}{sl}
\SetMathAlphabet{\mathsfit}{bold}{\encodingdefault}{\sfdefault}{bx}{n}

\definecolor{textgray}{HTML}{6E6E73}
\usetikzlibrary{positioning, calc}
\usetikzlibrary{decorations.pathmorphing}

\makeatletter
\patchcmd{\wrong@fontshape}{\@gobbletwo}{}{}{}
\makeatother
\numberwithin{equation}{section} 
\tcbuselibrary{minted}
\setminted[python]{
  linenos,
  breaklines,
  fontsize=\footnotesize,
  xleftmargin=2em
}

\makeatletter
\AtBeginDocument{
  \urlstyle{sf}
  
}
\makeatother

\definecolor{light}{RGB}{125, 125, 125}
\crefname{tcb@cnt@pbox}{code}{code}
\Crefname{tcb@cnt@pbox}{Code}{Code}
\crefname{assumption}{assumption}{assumption}
\Crefname{assumption}{Assumption}{Assumptions}

\newtcolorbox[auto counter]{pbox}[2][]{
  colback=white,
  title=Code~\thetcbcounter: #2,
  #1,fonttitle=\sffamily,
  fontupper=\sffamily,
  arc=2pt,
  colframe=bgcolor,
  coltitle=fgcolor,
  colbacktitle=bgcolor,
  toptitle=0.25cm,
  bottomtitle=0.125cm
}

\makeatletter
\newcommand\applefootnote[1]{%
  \begingroup
  \renewcommand\thefootnote{}%
  \renewcommand\@makefntext[1]{\noindent##1}%
  \footnote{#1}%
  \addtocounter{footnote}{-1}%
  \endgroup
}
\makeatother

\definecolor{cverbbg}{gray}{0.90}

\title{Omni-Router: Sharing Routing Decisions in Sparse Mixture-of-Experts for Speech Recognition}

\author[*]{Zijin Gu}
\author{Tatiana Likhomanenko}
\author{Navdeep Jaitly}

\affiliation{Apple}

\abstract{
Mixture-of-experts (MoE) architectures have expanded from language modeling to automatic speech recognition (ASR).
Traditional MoE methods, such as the Switch Transformer, route experts independently within each layer.
Our analysis reveals that routers in most layers make expert choices that are not strongly correlated with the choices of the routers in other layers.
To increase the cooperation between experts in different layers and encourage greater specialization, we use a shared router across different MoE layers. We call this model \emph{Omni-router Transformer}.
Extensive experiments on a large-scale pseudo-labeled dataset and evaluations across 10 diverse, out-of-domain ASR benchmarks demonstrate that the Omni-router Transformer is able to achieve lower training loss and consistently outperform dense and Switch Transformer models, reducing average word error rates by 11.2\% and 8.2\%, respectively, while providing structured expert usage and improved robustness to diverse data.}

\metadata[Code]{\url{{https://github.com/apple/ml-omni-router-moe-asr}}}
\metadata[Correspondence]{\sffamily Zijin Gu: \url{zijin@apple.com}}
\date{\sffamily\today}

\begin{document}

\maketitle

\section{Introduction}
Automatic speech recognition (ASR) systems have made remarkable progress in recent years, driven by advances in neural network architectures and the availability of large-scale datasets~\citep{vaswani2017attention,radford2023robust,gulati2020conformer,kahn2020libri}. Despite these advancements, achieving high recognition accuracy across diverse acoustic and linguistic conditions remains a challenging problem, particularly given the variability in speaker accents, background noise, and domain-specific vocabularies. 

There are several ways to improve ASR performance. Combining an acoustic model with additional language models (LMs), such as $n$-gram LMs~\citep{amodei2016deep,synnaeve2020end}, neural LMs~\citep{amodei2016deep,synnaeve2020end,zeyer18_interspeech,zeghidour2018fully} and error correction LMs~\citep{guo2019spelling,gu2024denoising}, were found to be effective ways to produce better transcriptions. Other approaches are inspired by the development of the scaling laws~\citep{brown2020language,hoffmann2022training,chen2025owls}, where scaling up model and data size is crucial to improve model performance. However, classical large models often face trade-offs between computational efficiency and representational power, especially when deployed in real-world scenarios where resource constraints and robustness are critical considerations~\citep{he2024upcycling,dai2024deepseekmoe}.

Mixture-of-experts (MoE) models~\citep{shazeer2017outrageously} have emerged as a promising architecture for their ability to scale model capacity (in terms of number of parameters) without incurring proportional overhead at inference. Unlike traditional dense models, which activate all parameters for every input, MoE models dynamically route input data to a subset of specialized experts tailored to process particular characteristics of the data. 
This routing mechanism enables MoE models to scale to significantly larger parameter counts without proportional increases in inference cost, offering both efficiency and flexibility. 
ASR involves highly diverse speech signals, with differences in pronunciation, accent, and noise patterns and MoE's, by design, can be quite appropriate to modeling such diversity.
By assigning specific experts to different aspects of the input, MoE models could better capture this variability, potentially improving recognition performance and robustness. Moreover, their ability to adaptively allocate resources makes them well-suited for deployment in heterogeneous environments, from edge devices with constrained resources to cloud-based systems with ample inference capacity.
Recently, \citep{jelassi2024mixture} showed that increasing the number of experts (while fixing the number of active parameters) increases the memorization performance consistently while the reasoning capabilities saturate;~\citep{abnar2025parameters} showed interaction between compute per example
and total number of parameters, and their impact on model in MoEs: there is an optimal level of sparsity that improves both training efficiency and model performance.

\begin{figure}[t]
  \centering
  \includegraphics[width=0.5\linewidth]{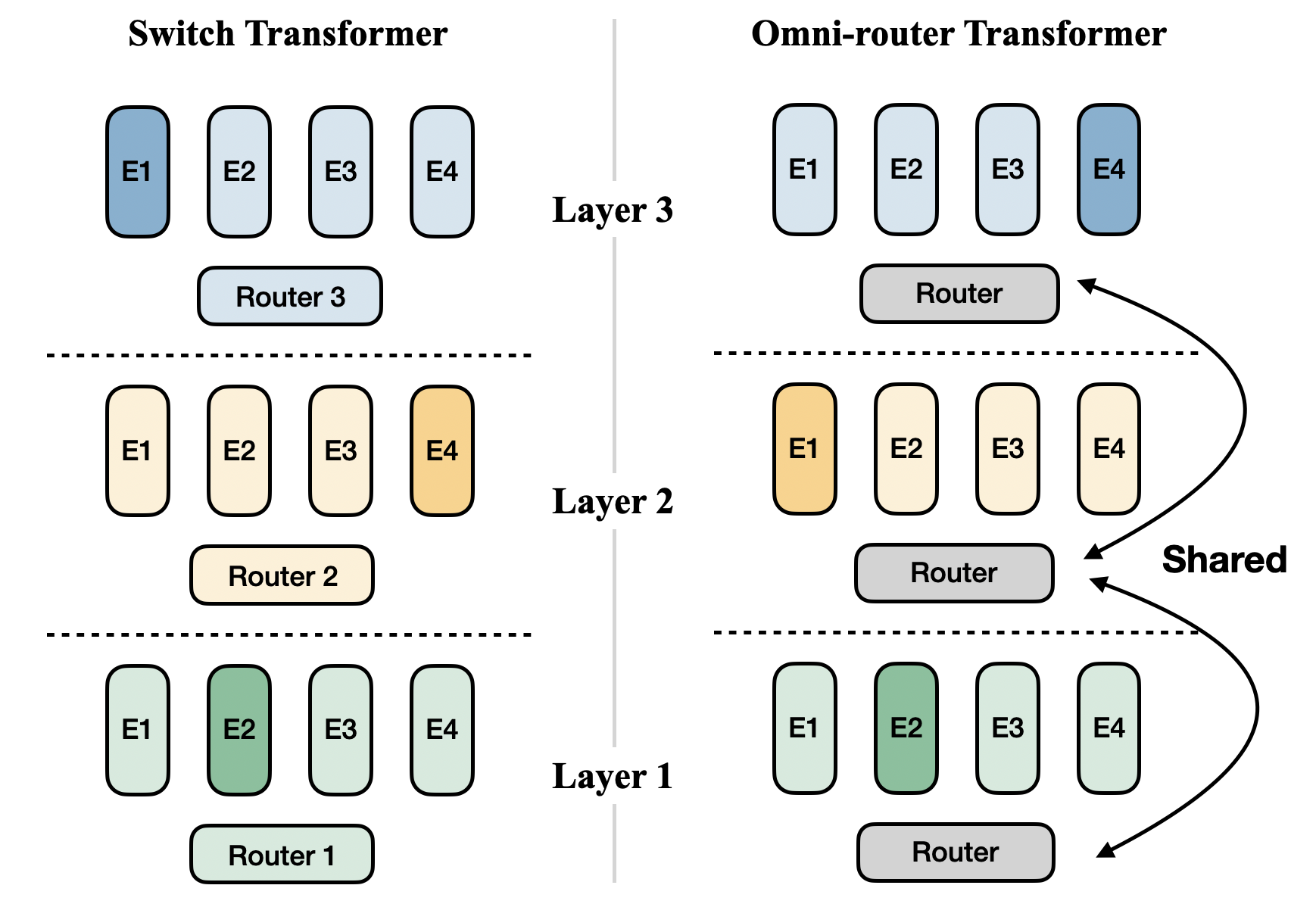}
  \caption{Illustration of Switch (left) and Omni-router (right) Transformers. $E_i$~represents an expert.}
  \label{fig:diagram}
\end{figure}

The effectiveness of MoE architectures heavily relies on the routing mechanism that directs inputs to appropriate experts.
Conventional MoE methods learn both the routers and expert parameters at the same time, with stochastic gradient descent (SGD). The learning is complicated by issues such as uneven expert utilization~\citep{zhou2022mixture}, insufficient expert specialization~\citep{dai2024deepseekmoe}, and high routing computational costs~\citep{dai2022stablemoe} often resulting in training instability~\citep{zoph2022st}.
While these issues are well explored, in this paper we highlight an under-explored issue - namely that the correlation between which experts are used at different layers is weak, with different layers making seemly arbitrary, independent decisions about which experts to use.
We hypothesize that this would lead to models that do not specialize to data very strongly.
To address this issue we propose to use a shared router across all layers that encourages coordinated decision-making across MoE layers (see architecture in Fig.~\ref{fig:diagram}).
We think it is possible to have a shared router across different layers, because of the residual nature of transformers - features from each layer are modified by residual connections and only change minimally from layer to layer~\citep{painters}, and thus sharing a decision boundaries across layers is possible by just using a shared router. 
Empirical results across multiple model configurations consistently indicate that Omni-router delivers superior lower training error, test word error rates (WER) and robustness against highly diverse data compared to traditional dense models and Switch Transformer baselines across diverse out-of-domain evaluation datasets.

\section{Related work}
MoE models were first introduced by Jacobs et al.~\citep{jacobs1991adaptive} as a dynamic mechanism to combine multiple specialized sub-models through gating functions, with subsequent hierarchical extensions presented by Jordan and Jacobs~\citep{jordan1994hierarchical}. Modern advances have significantly scaled MoE architectures to efficiently accommodate extremely large model sizes, exemplified by sparsely gated MoE layers proposed by Shazeer et al~\citep{shazeer2017outrageously}. Further efficiency improvements were realized with the Switch Transformer~\citep{fedus2022switch}, simplifying the routing mechanism by activating only one expert per input token. Additional strategies, such as BASE layers~\citep{lewis2021base} and hash-based routing techniques~\citep{roller2021hash}, have been explored to mitigate issues related to load imbalance and training inefficiency. MoE architectures have demonstrated substantial success across various domains, including natural language modeling (e.g., GLAM~\citep{du2022glam}) and computer vision tasks~\citep{riquelme2021scaling}.

In contrast, MoE applications within ASR remain comparatively limited. SpeechMoE~\citep{you2021speechmoe} proposes a shared embedding network and concatenates the shared embedding with output of the previous layer as the input of routers, meanwhile adding more loss terms, such as $L_1$, mean importance and embedding losses. Building on this work, SpeechMoE2~\citep{you2022speechmoe2} proposes a new router architecture by integrating additional global domain and accent embedding into router input to promote adaptability. Both SpeechMoE and SpeechMoE2 predominantly focus on monolingual speech recognition tasks. 

Recently, the application of MoE architectures to multilingual ASR has garnered increasing attention~\citep{hu2023mixture, kwon2023mole, li2024monet}. Hu et al~\citep{hu2023mixture} integrate MoE layers within conformer architectures~\citep{gulati2020conformer}, demonstrating an average relative improvement in WER of 11.9\% across twelve languages. Kwon et al.~\citep{kwon2023mole} propose a MoE model consisting of language-specific experts alongside a language-agnostic expert, while Li et al.~\citep{li2024monet} utilize routers that select optimal expert subsets combined with joint Connectionist Temporal Classification (CTC)/Attention training~\citep{kim2017joint} and rescoring, achieving substantial improvements in low-resource multilingual ASR scenarios.

Despite these advancements, critical challenges persist, particularly concerning load balancing, model stability, and training robustness~\citep{zoph2022st}. Such issues motivate the exploration of novel routing mechanisms aimed at enhancing MoE model robustness and effectiveness. Our contributions address these gaps in two primary ways. First, we introduce an innovative shared-routing approach within Transformer-based MoE ASR models, demonstrating its effectiveness in terms of both accuracy and robustness compared to conventional dense and Switch Transformer MoE architectures. Second, our proposed MoE design remains straightforward, eliminating the need for auxiliary embedding networks, complex auxiliary loss combinations~\citep{you2021speechmoe, you2022speechmoe2}, or explicit inductive biases such as accent or language embeddings~\citep{you2022speechmoe2, kwon2023mole}, thus facilitating simpler implementation and training while being effective.

\begin{figure*}[t]
  \centering
  \includegraphics[width=0.8\linewidth]{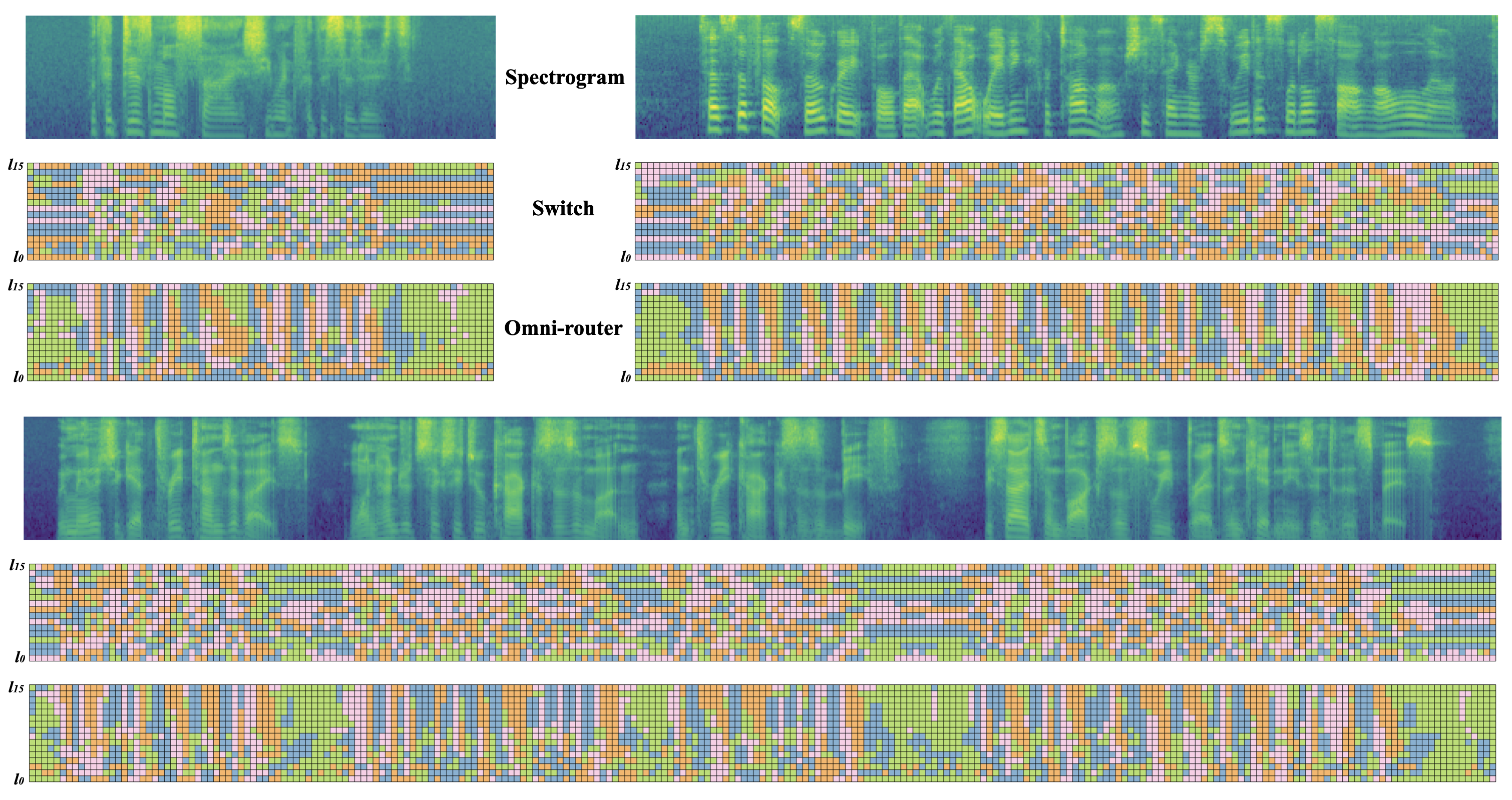}
  \caption{Visualization of randomly selected audio samples from the Librispeech \textit{dev-other} dataset, showing (top) their corresponding mel-spectrograms; (middle) expert usage patterns in a four-expert (blue, orange, green, pink) Switch Transformer ASR model; and (bottom) expert usage patterns in a four-expert Omni-router Transformer ASR model. Expert assignments are displayed across transformer layers ($l_i$, y-axis) and time ($t$, x-axis).}
  \label{fig:expert_usage}
\end{figure*}

\section{Background}

\subsection{Mixture-of-Experts layer}
Most works consider MoE in replacement of the feed-forward network (FFN) layer in transformer models~\citep{vaswani2017attention}. Typically, an MoE layer contains a routing network and a branch of experts, which are also FFNs. The router dispatches the incoming $T$~tokens $X^l\in\mathbb{R}^{T\times D}$ at layer $l$ to different experts first by estimating the weights $P\in\mathbb{R}^{T\times N}$ for each of $N$ experts:
\begin{align}
  P^l(X^l) = \mathrm{Softmax}(X^l W^l), \qquad l=1,...,L,
\end{align}
where $W^l\in\mathbb{R}^{D\times N}$ is the weight matrix of the router in a layer~$l$, and softmax is taken along the second dimension.
The top-$k$ experts of a specific token $X^l_i\in\mathbb{R}^{D}$ are then selected for routing it to give output $Y_i^l\in\mathbb{R}^{T\times D'}$:
\begin{align}
    Y^l_i = \sum_{j \in \mathcal{K}} [P^l(X^l)]_i^j \,\, [E_{j}(X^l)]_i,
\end{align}
where $\mathcal{K}$ is a set of indices of top-$k$ experts, and $E_{j}(\cdot)\in\mathbb{R}^{T\times D'}$ represents the output of expert $j$. 

\subsection{Switch Transformer}
Instead of choosing top-$k$ experts based on the router's decision, Switch Transformer~\citep{fedus2022switch} simplifies the routing mechanism and chooses only the top-1 expert. It was shown that switch layer not only reduces the router computation cost but also gives better performance in language models. Throughout this paper, we also use only the top-1 expert.

\subsection{Load balancing loss}
During training of MoE models, an auxiliary loss is often introduced to  maintain the balance of load (the number of tokens that go through one expert) of each expert. As introduced in~\citep{fedus2022switch}, given $N$ experts and $T$ tokens $X^l$ for layer $l$, the fraction of tokens dispatched to each expert $j$, $f^l$, and the fraction of router probability allocated for each expert $j$, $\rho^l$, are defined as
\begin{align}
    f_j^l = \frac{1}{T} \sum_{i=1}^T \mathbbm{1}\{\arg\max_{j'} [P^l(X^l)]_i^{j'} = j\}, \\
    \rho^l_j = \frac{1}{T}\sum_{i: \arg\max_{j'} [P^l(X^l)]_i^{j'} = j}[P^l(X^l)]^j_i, 
\end{align}
the load balancing loss is defined as  $\mathcal{L}_{\mathrm{load}}=\sum_{l=1}^L\mathcal{L}^l_{\mathrm{load}}$, where each $\mathcal{L}^l_{\mathrm{load}}$ is a scaled dot product:
\begin{align}
    \mathcal{L}^l_{\mathrm{load}} = N \sum_{j=1}^{N} f_j^l \cdot \rho_j^l.
\end{align}

\section{Omni-router: sharing routing decisions across MoE layers}
In conventional MoE architectures, such as the Switch Transformer, each MoE layer independently selects its experts without explicit coordination among layers. Although SGD facilitates implicit coordination between these layers during training, the inherent randomness introduced by SGD can lead the model to function more as an ensemble of loosely coupled experts rather than as a tightly integrated system. This phenomenon is analogous to the effect observed in dropout training, where the model adapts to arbitrary subsets of active parameters.

While learning uncorrelated experts can be beneficial in low-data regimes by reducing overfitting, this property may not be advantageous in high-data settings where the emergence of highly specialized experts is desirable. We argue that, for such strong specialization to occur, different layers of a transformer must collaboratively establish shared decision boundaries around input data points. To achieve this, we couple the routing parameters across transformer layers by sharing the router weights across different MoE layers: instead of having a $W^l$ for $l$-th layer, we have a $W^{\mathrm{shared}}$ for all layers:
\begin{align}
  P^l(X^l) = \mathrm{Softmax} (X^lW^{\mathrm{shared}}).
\end{align}

This strategy is viable due to the prevalent use of pre-layer normalization in transformer architectures, which incorporates residual connections directly linking input and output representations. Consequently, feature representations between consecutive layers remain sufficiently similar, enabling shared routing parameters to establish coherent—though not identical—decision boundaries across layers. Meanwhile, the dense parameters within each MoE layer remain free to specialize independently, adapting to layer-specific decision boundaries. This approach contrasts with existing parameter-sharing methods in MoE architectures, which typically involve sharing the experts' weights rather than the routing parameters themselves~\citep{csordas2024moeut,zhang2024efficient,zhao2024hypermoe}.

\section{Empirical setup}
\subsection{Data}
We utilize a large-scale conversational audio dataset collected from publicly accessible sources, named SpeechCrawl. This dataset was specifically curated to be diverse, conversational, multilingual, and multi-speaker. Audio samples average approximately 30 minutes in duration, with around 60\% consisting of English speech. Since SpeechCrawl does not provide transcriptions, we generate pseudo-labels using the WhisperX~\citep{bain2023whisperx} pipeline in combination with Whisper's large-v2 multilingual model~\citep{radford2023robust}. These pseudo-labels simultaneously yield audio segmentation (with segments shorter than 30 seconds) and corresponding transcripts.
To ensure quality, we filter out unreliable pseudo-labels resulting from looping artifacts of decoding in the sequence-to-sequence model. Specifically, we employ a simple heuristic based on expected speech rates (approximately three words per second), similar to~\citep{likhomanenko2021slimipl}. After filtering, we retain approximately 1M hours of English speech segments for training.
Additionally, we randomly select about 12h of audio (24 files) as our development set to monitor training progress and identify the optimal checkpoint for subsequent evaluation on out-of-domain (OOD) datasets. For OOD testing, we employ diverse test sets from AMI-IHM~\citep{carletta2007unleashing}, Callhome~\citep{godfrey1992switchboard}, Chime6~\citep{watanabe2020chime}, CommonVoice~\citep{ardila2019common}, Fleurs~\citep{conneau2023fleurs}, Librispeech~\citep{panayotov2015librispeech}, Switchboard~\citep{godfrey1992switchboard}, Tedlium~\citep{hernandez2018ted}, Voxpopuli~\citep{wang2021voxpopuli}, and Wall Street Journal~\citep{paul1992design}. \textit{Importantly, none of the selected test datasets overlap with the training data.}

\subsection{ASR models}
We compare our proposed Omni-router Transformer model against two baseline architectures: a dense transformer and a Switch Transformer. All ASR models utilize the standard Transformer encoder architecture~\citep{vaswani2017attention}, optimized using CTC~\citep{graves2012connectionist} and employing an 8k-token word-piece vocabulary derived from the training transcriptions.

\textbf{Baseline 1: Dense Transformer}~~~
We train three dense transformer models of varying sizes—84M, 140M, and 200M parameters—with corresponding embedding dimensions of 512, 768, and 1024, respectively. All dense models consist of 16 encoder blocks, each with a FFN dimension of 4096.

\textbf{Baseline 2: Switch Transformer}~~~
Corresponding to each dense model, we train a Switch Transformer model by replacing the FFN layers with MoE layers. Each expert within the MoE layers retains the same FFN structure as the corresponding dense baseline. During both training and inference, each layer dynamically selects the top-1 expert based on the router's decision.

\textbf{Proposed Model: Omni-router Transformer}~~~
For each dense model configuration, we train an Omni-router Transformer model, introducing a shared router across all layers to encourage consistent token routing decisions. Apart from this shared-routing mechanism, the Omni-router Transformer maintains the same configuration as the Switch Transformer baseline.

\subsection{Training setup}
All ASR models take as input 80-dimensional log-mel filterbank features computed using a 25ms sliding window with a 10ms stride. Subsequently, every four consecutive frames are stacked before being fed into the transformer layers. We apply SpecAugment~\citep{park2019specaugment} as data augmentation, employing 2 frequency masks (maximum width 30) and 10 time masks (maximum width 50, ratio 0.1). Training batches dynamically combine audio segments up to a total of 7 hours, with gradient clipping set at 0.1. The learning rate (LR) schedule comprises a warm-up phase of 64k steps reaching a peak LR of 0.001, followed by cosine decay for the next 60k steps, and a step decay at a factor of 0.5 thereafter. For MoE ASR models, we include an auxiliary load balancing loss $\mathcal{L}_{\mathrm{load}}$ to encourage even expert utilization, assigning it a weight of 10 based on its relative scale compared to the CTC loss. All models are optimized using AdamW with a weight decay of 0.01 and trained for up to 1M steps (7 epochs) on 16 H100 GPUs, terminating once the greedy validation word error rate (WER) plateaus. 
Hyperparameters were first optimized using the baseline dense ASR model (84M parameters) and subsequently held constant for all dense and MoE variants. This approach was motivated by the higher computational cost of tuning MoE models. By tuning the dense model and scaling to MoE, we aimed to validate the generalization capabilities of MoE architectures while enabling resource-efficient optimization. 
However, for some Switch Transformer models, we reduced the peak LR and adjusted the load balancing loss weight to address observed training instability.

\subsection{Evaluation}
We evaluate the final model performance by computing word error rates (WER) using greedy decoding. Before calculating WER, we apply the Whisper normalizer described in~\citep{radford2023robust} to both ground truth and predicted transcriptions.

\section{Results}
\subsection{Omni-router improves ASR}

In Table~\ref{tab:main}, we present the WER results comparing a baseline dense ASR model (140M) with Switch Transformer and Omni-router Transformer ASR models, each employing two experts per layer. 
We first note that both MoE-based architectures consistently outperform the dense baseline across all evaluated out-of-domain datasets. Additionally, the Omni-router Transformer further improves upon the performance of the Switch Transformer. Given that the Omni-router shares hyperparameters originally optimized for the Switch Transformer, its superior performance despite minimal hyperparameter tuning underscores the effectiveness and robustness of the shared-routing strategy.

\begin{table}[h]
  \caption{WERs of a dense ASR, a 2-expert Switch Transformer ASR and a 2-expert Omni-router Transformer ASR for different OOD test datasets.}
  \label{tab:main}
  \centering  
  \setlength{\tabcolsep}{6pt}
  \renewcommand{\arraystretch}{1.3}
  \begin{tabular}{c|r|r|r}
    \hline
    \textbf{Dataset} & \textbf{Dense} & \textbf{Switch} & \textbf{Omni-router} \\
    \cline{1-4} 
    SpeechCrawl & 6.1 & 5.8 & \textbf{5.4} \\
    \hline
    AMI-IHM & 18.5 & 18.3 & \textbf{17.8}\\
    \hline
    Chime6 & 29.3 & 29.4 & \textbf{28.2} \\
    \hline
    CommonVoice & 19.6 & 18.9 & \textbf{16.4}\\
    \hline
    Fleurs & 9.9 & 9.0 & \textbf{8.4}\\
    \hline
    Callhome & 15.3 & 15.5 & \textbf{15.0} \\
    \hline
    Switchboard & 14.1 & 14.1 & \textbf{13.5} \\
    \hline
    Librispeech (clean) & 4.2 & 3.9 & \textbf{3.3} \\
    \hline
    Librispeech (other) & 9.0 & 8.4 & \textbf{7.3} \\
    \hline
    WSJ (nov92) & 4.4 & 4.2 & \textbf{3.7}\\
    \hline
    Tedlium & 4.7 & 4.6 & \textbf{4.2}\\
    \hline
    Voxpopuli & 9.5 & 9.1 & \textbf{8.5}\\
    \hline
  \end{tabular}
\end{table}

\subsection{Why Does the Omni-router Transformer Outperform}

A central question raised by our experimental results is what factors contribute to the superior performance of the Omni-router Transformer. We hypothesize that the shared routing mechanism creates stable and structured "pathways" for tokens through the network, enabling the model to develop more cooperative and specialized experts.

To investigate this hypothesis, we conducted a detailed analysis of expert utilization, as illustrated in Fig.~\ref{fig:expert_usage}. We visualize expert assignments across three randomly selected audio samples from the Librispeech \textit{dev-other} dataset. Each sample visualization consists of (from top to bottom): the mel-spectrogram, the expert assignments for a 4-expert Switch Transformer, and the expert assignments for a 4-expert Omni-router Transformer. Expert assignments are plotted with time steps on the horizontal axis and transformer layers on the vertical axis. The colors representing experts have no inherent meaning, and for clarity in Switch Transformer plots, colors are permuted to optimally align across layers.

Our visualizations reveal significant differences in routing behavior between the two architectures. The Omni-router Transformer exhibits notably structured and coherent expert assignments across both temporal and depth dimensions, whereas the Switch Transformer demonstrates a more fragmented and less organized routing pattern. Specifically, the Omni-router’s routing creates consistent temporal segments, suggesting meaningful coordination and communication between layers. Furthermore, repeated utilization of the same experts across multiple layers indicates that the Omni-router’s inductive bias effectively promotes stable specialization. Crucially, expert assignments in the Omni-router Transformer clearly correlate with acoustic patterns in the mel-spectrogram. For example, the "green" expert predominantly handles silent segments, while alternating patterns of orange, blue, and pink experts correspond clearly to distinct speech segments. In contrast, the Switch Transformer displays fragmented expert assignments, indicating weaker inter-layer coordination and suboptimal specialization.

\begin{figure}[h]
  \centering
  \includegraphics[width=0.6\linewidth]{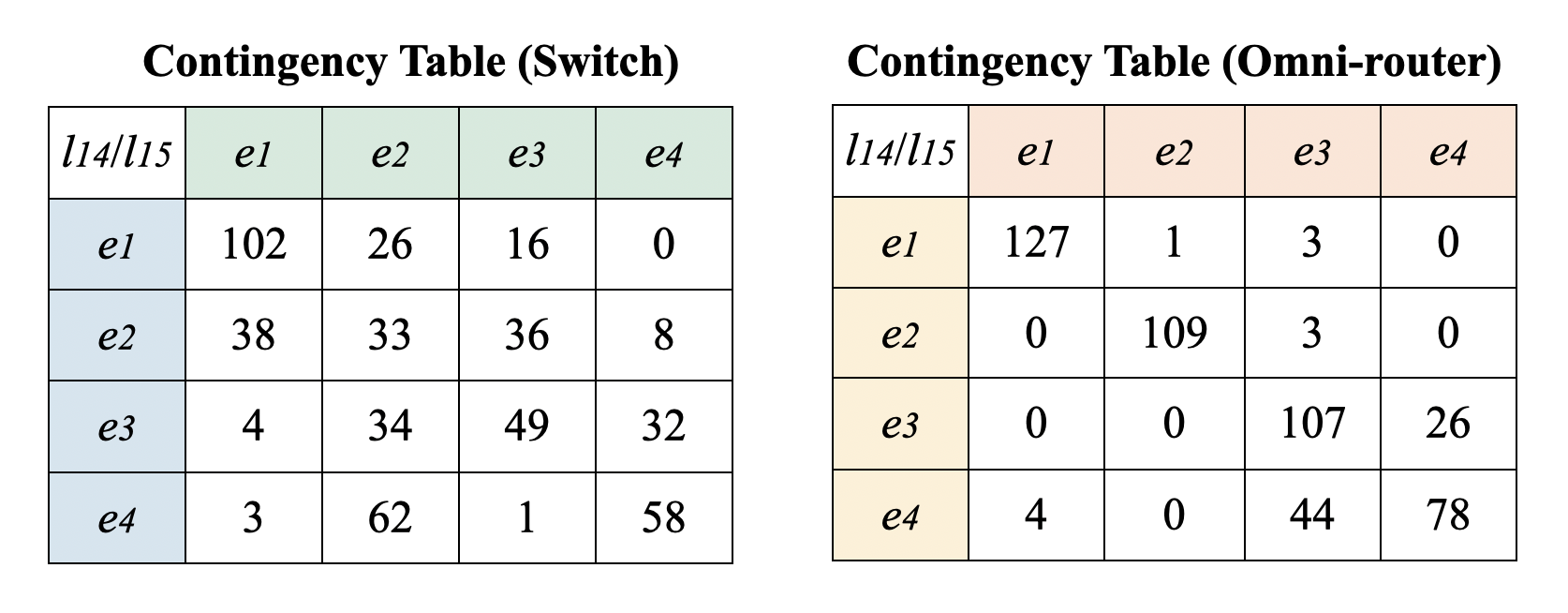}
  \caption{Contingency tables illustrate the expert correspondence between layers 14 and 15 for both Switch Transformer and Omni-router Transformer as representative examples.}
  \label{fig:contigency}
\end{figure}

\begin{figure}[h]
  \centering
\includegraphics[width=0.6\linewidth]{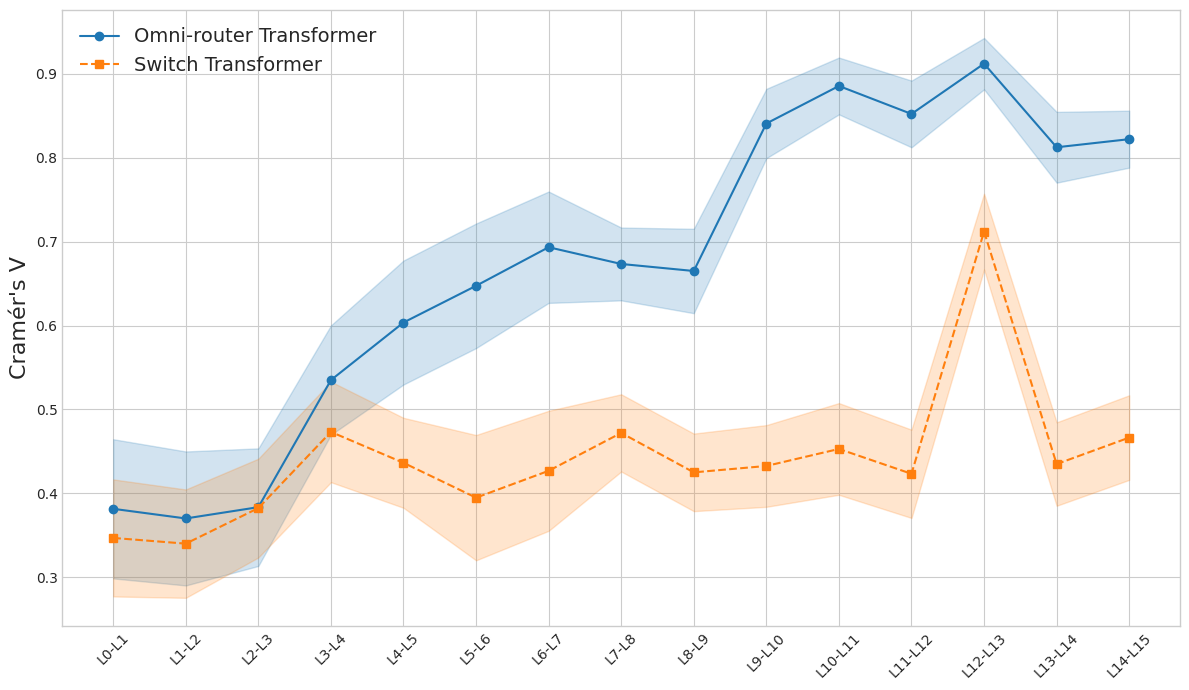}
  \caption{Expert correlation (Cramér's V) between adjacent layers in Switch Transformer and Omni-router Transformer.}
  \label{fig:corr}
\end{figure}

To quantitatively substantiate these qualitative observations, we conducted analyses along three dimensions using the Librispeech \textit{dev-other} dataset:

\textbf{Enhanced Inter-Layer Cooperation}: To understand how expert assignments evolve through layers, we examined the consistency of expert choices between adjacent layers. Fig.~\ref{fig:contigency} depicts this relationship using contingency tables for layers 14 and 15, with expert indices reordered for optimal alignment. Extending this analysis network-wide, Fig.~\ref{fig:corr} shows Cramér’s V correlation values computed across all adjacent layer pairs. The results indicate that the Omni-router Transformer consistently achieves higher inter-layer expert assignment correlation, particularly pronounced in deeper network layers, reinforcing its stronger inter-layer cooperation.

\textbf{Lower Expert Entropy}: We further analyzed the confidence in routing decisions by evaluating expert entropy distributions across layers for the top 100 most frequently occurring tokens. The Omni-router Transformer consistently exhibits lower entropy compared to the Switch Transformer, indicating higher confidence and decisiveness in expert selection.

\begin{figure}[h]
  \centering
\includegraphics[width=0.4\linewidth]{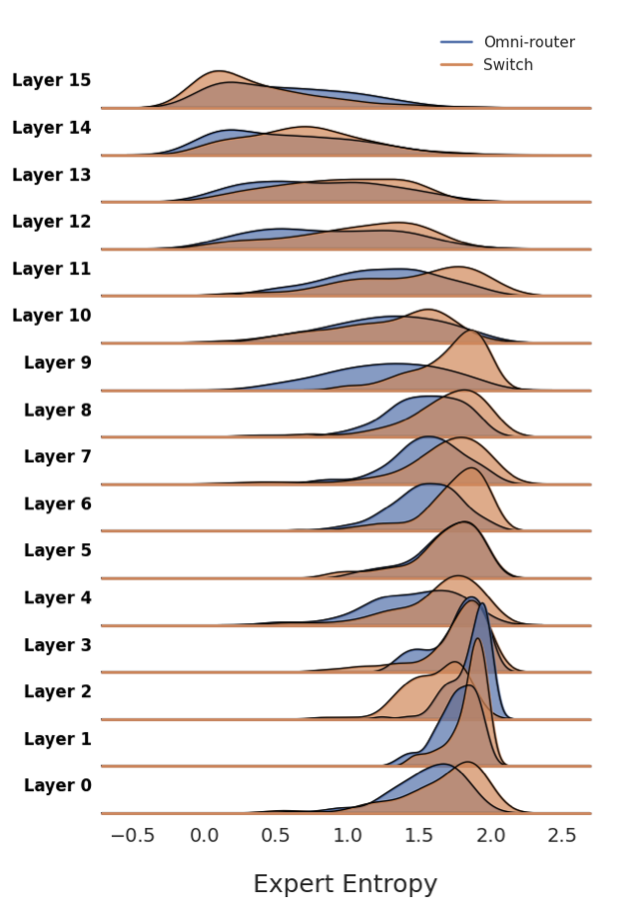}
  \caption{Expert entropy across layers of Switch Transformer and Omni-router Transformer.}
  \label{fig:entropy}
\end{figure} 

\textbf{Greater Expert Specialization}: To rigorously quantify expert specialization, we adopted an expert permutation method proposed in prior research \citep{dai2024deepseekmoe}. Specifically, we randomly permuted expert assignments for each token with varying probability $p$ and measured the resulting impact on ASR performance, quantified by WER degradation. We hypothesized that the Omni-router Transformer would exhibit a greater sensitivity to incorrect expert assignments, reflecting higher expert specialization.

\begin{figure}[h]
  \centering
\includegraphics[width=0.6\linewidth]{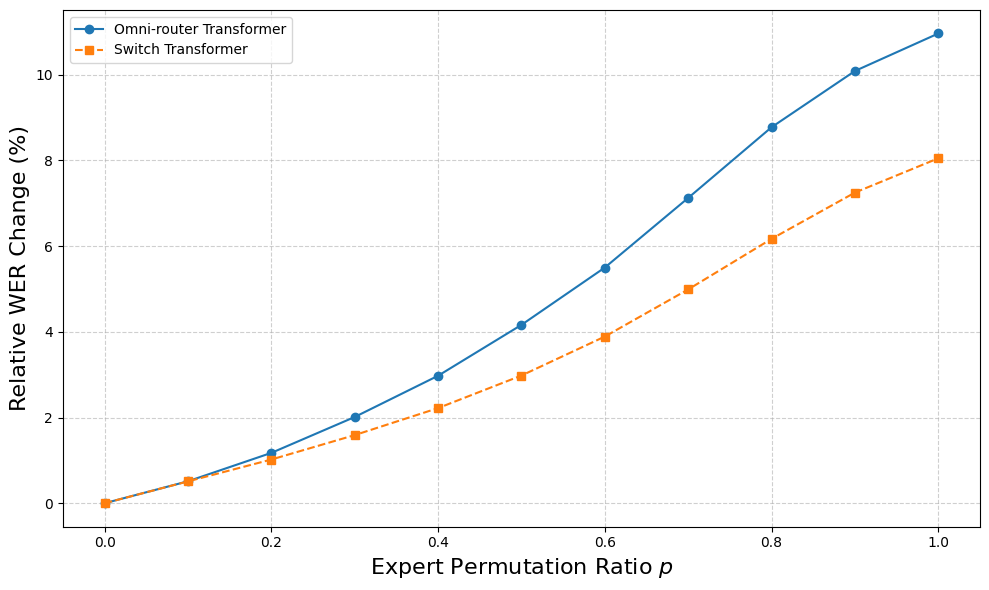}
  \caption{Relative WER change percentage of Switch Transformer ASR and Omni-router Transformer ASR at different permutation ratios $p$.}
  \label{fig:specialization}
\end{figure}

Fig.~\ref{fig:specialization} supports our hypothesis, clearly illustrating that the Omni-router Transformer experiences a significantly larger performance degradation (increased WER) as the permutation ratio $p$ increases, compared to the Switch Transformer. This indicates a stronger reliance on accurate, specialized expert assignments and reinforces the conclusion that the Omni-router Transformer promotes enhanced expert specialization and cooperative behavior across its architecture.

\subsection{Omni-router Transformer outperforms Switch Transformer across different number of experts}
In Table~\ref{tab:num_expert}, we investigate how varying the number of experts impacts the performance of Switch Transformer and Omni-router Transformer-based ASR models, compared against a baseline dense ASR model (84M parameters).
We present results on the Librispeech dataset (we observed similar trends across other datasets). Our findings demonstrate that Omni-router ASR consistently achieves superior performance over Switch Transformer ASR across all evaluated expert configurations. Notably, Switch Transformer's performance deteriorates at 8 experts, 
while Omni-router's structured layer-coupled routing effectively mitigates this degradation, maintaining stable performance even when using more  experts.

\begin{table}[th]
  \caption{WERs of Switch Transformer ASR and Omni-router Transformer ASR models with different number of experts.}
  \label{tab:num_expert}
  \centering
   \setlength{\tabcolsep}{6pt}
  \renewcommand{\arraystretch}{1.3}
  \begin{tabular}{c|c|c|r|r}
    \hline
    \textbf{Model} & \textbf{$n$ experts} &  \textbf{Model size} & \textbf{test-clean} & \textbf{test-other}\\
    \cline{1-5}    
    Dense & - & 84M & 5.3 & 10.9 \\
    \hline
    Switch & 2 & 156M &4.4 & 9.6 \\
    Omni-router & 2 & 156M& \textbf{4.2} & \textbf{9.0}\\
    \hline
    Switch & 4 & 290M &4.5 & 10.0 \\
    Omni-router & 4 & 290M & \textbf{3.7} & \textbf{7.9}\\
    \hline  
    Switch & 8 & 559M &8.6 & 17.6 \\
    Omni-router & 8 & 559M & \textbf{3.9} & \textbf{8.1} \\
    \hline
  \end{tabular}
\end{table}

\subsection{Omni-router Transformer outperforms Switch Transformer across different model sizes}
In Table~\ref{tab:model_size}, we examine the impact of model size by evaluating baseline dense ASR models with parameter counts of 84M, 140M, and 200M. Correspondingly, we train sets of 2-expert Switch Transformer and 2-expert Omni-router Transformer ASR models matched to these sizes. We find that the Omni-router ASR models consistently achieve lower WERs compared to both the Switch Transformer ASR models and the baseline dense ASR models.

\begin{table}[th]
  \caption{WERs of Dense models, and Switch Transformer and Omni-router Transformer ASR models with different number of model sizes and two experts.}
  \label{tab:model_size}
  \centering
    \setlength{\tabcolsep}{6pt}
  \renewcommand{\arraystretch}{1.3}
  \begin{tabular}{c|c|r|r}
    \hline
    \textbf{Model} & \textbf{Model size} & \textbf{test-clean} & \textbf{test-other}\\
    \cline{1-4}  
    Dense & 84M & 5.3 & 10.9 \\
    Switch & 156M  & 4.4 & 9.6 \\
    Omni-router& 156M  & \textbf{4.2} &  \textbf{9.0} \\
    \hline
    Dense & 140M & 4.1 & 9.0\\
    Switch & 246M & 3.9 & 8.4 \\
    Omni-router & 246M &  \textbf{3.3} &  \textbf{7.3}\\
    \hline
    Dense & 200M & 3.7 & 8.0\\
    Switch & 345M & 4.2 & 8.7 \\
    Omni-router & 345M &  \textbf{3.4} &  \textbf{7.3}\\
    \hline
  \end{tabular}
\end{table}

\subsection{Results are reproducible on single and specialized datasets}
To validate the generalizability and robustness of the Omni-router Transformer beyond conversational data, we extended our evaluation to the domain of read speech. For this, we conducted experiments on the Libriheavy dataset~\citep{kang2024libriheavy}, a large-scale ASR corpus derived from audiobooks in the LibriVox project~\footnote{https://librivox.org}. We trained three models, each with approximately 84M active parameters: a dense baseline, an 8-expert Switch Transformer, and an 8-expert Omni-router Transformer. The models were trained on the Libriheavy training set, and evaluated on Librispeech \textit{dev} and \textit{test} splits.

\begin{table}[th]
  \caption{WERs of a dense ASR model, a Switch Transformer ASR model and a Omni-router Transformer ASR model trained on Libriheavy and evaluated on Librispeech dataset.}
  \label{tab:ls}
  \centering
    \setlength{\tabcolsep}{6pt}
  \renewcommand{\arraystretch}{1.4}
  \begin{tabular}{c|r|r|r}
    \hline
    \textbf{Dataset} & \textbf{Dense} & \textbf{Switch} & \textbf{Omni-router} \\
    \cline{1-4} 
    dev-clean & 2.1 & 1.9 & \textbf{1.8} \\
    \hline
    dev-other & 6.6  & 6.1 & \textbf{5.3} \\
    \hline
    test-clean & 2.3 & 2.1 & \textbf{2} \\
    \hline
    test-other & 6.1  & 5.7 & \textbf{5.2} \\
    \hline
  \end{tabular}
\end{table}

As shown in Table~\ref{tab:ls}, the Omni-router Transformer consistently achieves the lowest WER across all evaluation sets. Particularly, on \textit{dev/test-other} splits, Omni-router Transformer achieves a relative WER reduction of 13.1\%/8.8\% over the Switch Transformer model and 19.7\%/14.7\% over the dense baseline.
These results confirm that Omni-router Transformer's routing mechanism can successfully translate to the distinct characteristics of read speech, which is a more robust and effective architecture for ASR tasks in diverse acoustic environments.

\subsection{Omni-router Transformer demonstrates better robustness}
\begin{figure}[h]
  \centering
\includegraphics[width=0.6\linewidth]{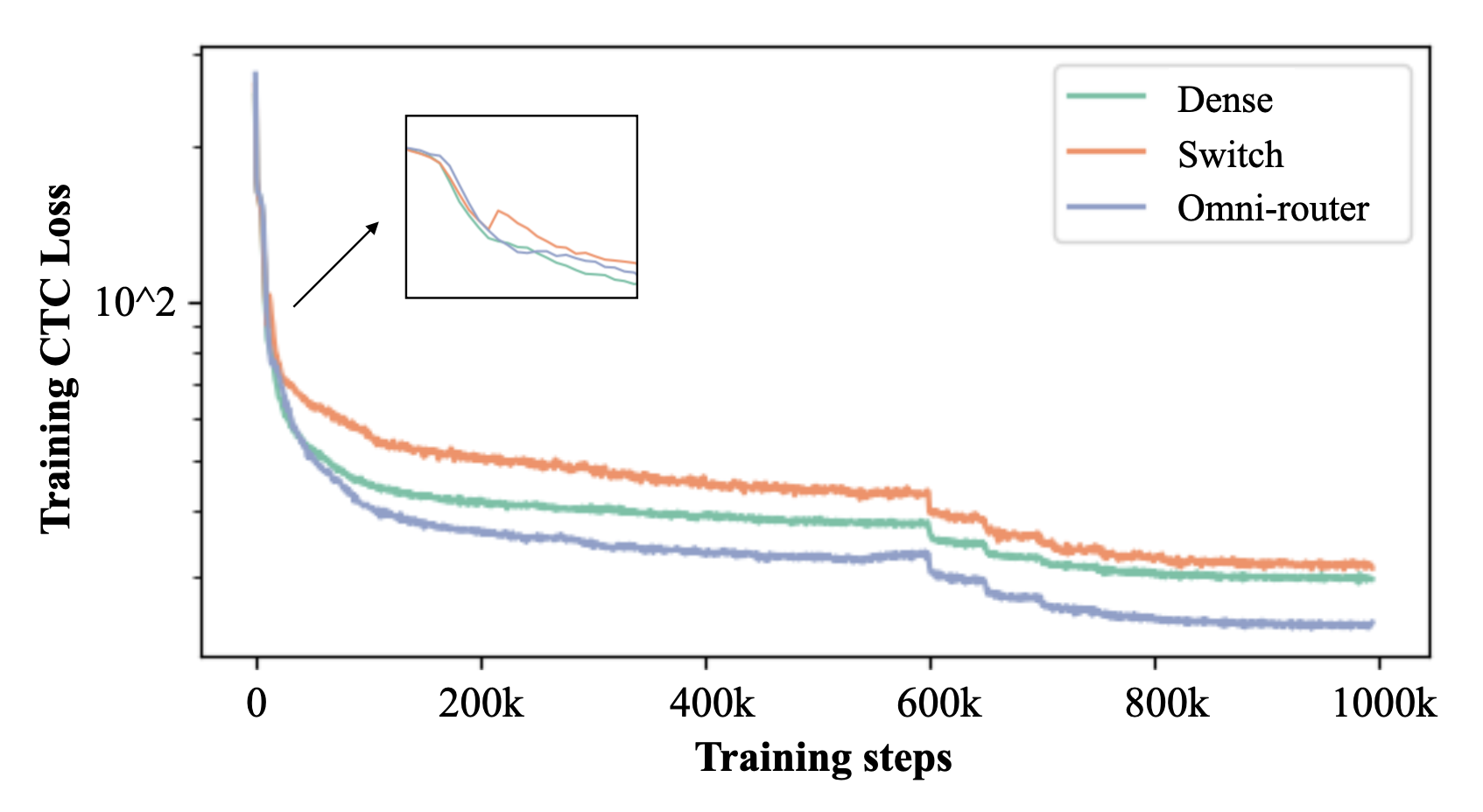}
  \caption{Training CTC loss of dense, Switch Transformer and Omni-router Transformer based ASR models on conversational data.}
  \label{fig:stability}
\end{figure}

In our experiments with SpeechCrawl data, we observed greater training instability in Switch Transformer models compared to Omni-router Transformer models, particularly as model size increased. This is illustrated in Fig.~\ref{fig:stability}, where the training CTC loss for the Switch Transformer briefly surpasses that of the dense baseline due to an early spike. Two potential contributing factors were identified: 1) the inherent noise in pseudo labels generated by WhisperX for SpeechCrawl data, and 2) the heterogeneous nature of conversational speech data.

To investigate these possibilities, we first relabeled Librispeech data using WhisperX following the same procedure applied to SpeechCrawl, subsequently training Switch Transformer ASR models. The resulting training loss remained stable, suggesting that the pseudo labels alone are unlikely to be the primary cause of instability. Next, we trained the model using real conversational datasets such as Fisher~\citep{cieri2004fisher}, observing training behaviors similar to those encountered with SpeechCrawl, including the characteristic early-stage spike. These findings indicate that data characteristics, rather than label quality, likely drive the observed instability. Conversely, Omni-router Transformer models consistently demonstrated stable training behavior across various data types, highlighting their robustness in large-scale ASR training scenarios.

\section{Comparisons with Dense Models}
To further showcase the effectiveness of Omni-router approach, we added a comparison of Omni-router ASR against dense baselines with similar total parameter counts. As shown in below table, the sparse Omni-router (246M total/140M active params) achieves a WER on par with both a dense model (210M) trained on the same data and the Whisper-small model (244M).

\begin{table}[h]
  \caption{WER comparisons between a sparse Omni-router ASR (246M total parameters with 140M active parameters) and dense models (210M and 244M). }
  \label{tab:whsiper}
  \centering  
  \setlength{\tabcolsep}{6pt}
  \renewcommand{\arraystretch}{1.3}
  \begin{tabular}{c|r|r|r}
    \hline
    \textbf{Dataset} &  \makecell{\textbf{Omni-router} \\ (246M - 140M)} & \makecell{\textbf{Dense} \\ (210M)} & \makecell{\textbf{Whisper-small.en} \\ (244M)} \\
    \cline{1-4} 
    AMI-IHM  & 17.8 & 17.8 & 17.6\\
    \hline
    Chime6 & 28.2 & 27.9 & 27.7 \\
    \hline
    CommonVoice & 16.4 & 17.6 & 15.2 \\
    \hline
    Fleurs & 8.4 & 8.8 & 7.6 \\
    \hline
    Callhome & 15.0 & 15.2 & 20.0 \\
    \hline
    Switchboard & 13.5 & 13.8 & 14.9 \\
    \hline
    \makecell{Librispeech \\ (clean)} & 3.3 & 3.7 & 3.1 \\
    \hline
    \makecell{Librispeech \\ (other)} & 7.3 & 8.0 & 7.4 \\
    \hline
    WSJ (nov92) & 3.7 & 4.1 & 3.5 \\
    \hline
    Tedlium & 4.2 & 4.3 & 4.0 \\
    \hline
    Voxpopuli & 8.5 & 9.0 & 8.2 \\
    \hline
  \end{tabular}
\end{table}

\section{Conclusion}
We introduce the Omni-router MoE architecture for speech recognition, which facilitates the sharing of routing decisions across layers, providing an inductive bias particularly well-suited to speech tasks. The Omni-router architecture is both simple and effective, demonstrating significant improvements in the efficiency and robustness of MoE model training. Experimental results consistently show that Omni-router-based ASR models achieve lower WERs compared to baseline dense models and Switch Transformer-based ASR models, across varying numbers of experts and model sizes. Furthermore, Omni-router models exhibit structured expert utilization with higher inter-layer correlation and more specialized experts. In addition, Omni-router models enhanced training robustness, notably showing increased tolerance to larger model sizes and number of experts. These findings highlight the critical importance of effective inter-layer coordination in MoE architectures and suggest promising avenues for further research into structured routing mechanisms to improve the accuracy and robustness of large-scale speech recognition systems.

\section*{Acknowledgment}
The authors thank Dan Busbridge, Jason Ramapuram, Russ Webb, and Shuangfei Zhai for their feedback.

% \applefootnote{ \textcolor{textgray}{\sffamily Apple and the Apple logo are trademarks of Apple Inc., registered in the U.S. and other countries and regions.}}

\newpage
\bibliographystyle{apalike}
\bibliography{mybib}

\end{document}